\documentclass{article}

\usepackage[preprint]{neurips_2021}

\usepackage[utf8]{inputenc} 
\usepackage[T1]{fontenc}    
\usepackage[colorlinks,
            linkcolor=red,
            anchorcolor=blue,
            citecolor=green,
            ]{hyperref}       
\usepackage{url}            
\usepackage{booktabs}       
\usepackage{amsfonts}       
\usepackage{nicefrac}       
\usepackage{microtype}      
\usepackage{xcolor}         
\usepackage{graphicx}
\usepackage{amstext}
\usepackage{amsmath}
\usepackage{algorithm}
\usepackage{algorithmic}
\usepackage{subfigure}
\usepackage{makecell}
\usepackage{multicol}
\usepackage{multirow}
\usepackage{array}
\usepackage{float}
\usepackage[sorting=none]{biblatex}
\usepackage{xr}
\usepackage{xr-hyper}
\newcommand{\fullmethodname}{Arch-Net: Model Distillation for Architecture Agnostic Model Deployment}
\newcommand{\methodname}{Arch-Net}

\title{\fullmethodname}

\author{Weixin Xu \quad Zipeng Feng \quad Shuangkang Fang \quad Song Yuan \\ {\bf \quad Yi Yang \thanks{Corresponding author.}\quad Shuchang Zhou} \\
Megvii Inc. \\
\{xuweixin02, fengzipeng, fangshuangkang, yuansong, yangyi, zsc\}@megvii.com}

\begin{document}

\maketitle

\begin{abstract}
  Vast requirement of computation power of Deep Neural Networks is a major hurdle to their real world applications. Many recent Application Specific Integrated Circuit (ASIC) chips feature dedicated hardware support for Neural Network Acceleration. However, as ASICs take multiple years to develop, they are inevitably out-paced by the latest development in Neural Architecture Research. For example, Transformer Networks do not have native support on many popular chips, and hence are difficult to deploy. In this paper, we propose \methodname, a family of Neural Networks made up of only operators efficiently supported across most architectures of ASICs. When a Arch-Net is produced, less common network constructs, like Layer Normalization and Embedding Layers, are eliminated in a progressive manner through label-free Blockwise Model Distillation, while performing sub-eight bit quantization at the same time to maximize performance. Empirical results on machine translation and image classification tasks confirm that we can transform latest developed Neural Architectures into fast running and as-accurate \methodname, ready for deployment on multiple mass-produced ASIC chips. The code will be available at \url{https://github.com/megvii-research/Arch-Net}.
\end{abstract}

\section{Introduction} \label{sec-introduction}
Deploying the computational intensive Deep Neural Networks(DNN) in real world scenarios has always been a challenge since their invention. Consequently, many recent Application Specific Integrated Circuit (ASIC) chips are dedicated to speeding up the inferences of DNNs. However, as ASICs take multiple years to develop, they are inevitably out-paced by the later progress in Neural Architecture research. For example, Transformer Networks emerging in Language Modeling and Computer Vision, do not enjoy native support on many popular chips, running considerably slower than expected. On the other hand, model quantization, especially the sub-eight-bit quantization capabilities of chips often remain under-explored, as the know-hows of quantizing Neural Networks diverge between chips. In this work, we propose Arch-Net, a family of Neural Networks built out of a small core set of almost-universally supported hardware operators. Arch-Net can be used to reduce the ever growing workload of supporting every Neural Architecture on every ASIC, by using a Blockwise Model Distillation method to morph the vastly diverse Neural Network Architectures into the simple family of Arch-Net. During the distillation, the bit width of weights and features can be reduced along the way thanks to utilities that accompanies Arch-Net. Empirical results on machine translation and image classification tasks confirm the efficacy of using Arch-Net as intermediate form between various Neural Architectures and the parade of Neural Network Accelerator ASICs. For example, we found that Layer Normalization and Embedding Layers in Transformers, can be transformed into more mundane Batch Normalization and Fully-Connected layers, while sustaining comparable accuracy.

To the Neural Architecture Designers, Arch-Net makes it possible to be \textit{agnostic} of Hardware Architecture while customizing Neural Architecture to exploit the inductive bias of data, as long as the designed Neural Architecture can be distilled into a Arch-Net.

To the Architects of Neural Network Accelerators, \methodname~provides a list of high priority operators making up Arch-Net. Supporting these operators make it possible to be \textit{agnostic} of Neural Architectures to some extent, and significantly boost the future-proof level of an ASIC in the face of ever evolving Neural Architectures.

\begin{figure}
    \centering
    \includegraphics[width=1.0\textwidth]{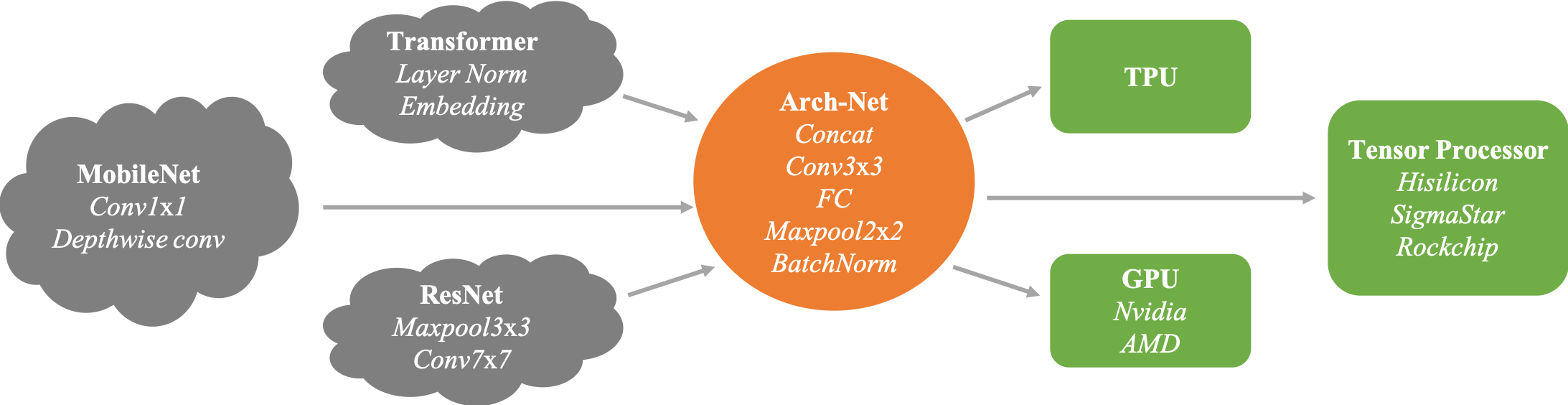}
    \caption{Overview of \methodname}
    \label{Fig-overview-archnet}
\end{figure}

Picking the core operator set of Arch-Net can be a challenging task given the plethora of Neural Architectures. Fortunately, the requirement of universal hardware support will make the candidate list very short in the first place. The question next is devising transformation rules to transform less common constructs into more mundane constructs. Thankfully, as the case with PReLU and LeakyRelu, many constructs do not differ during inference time. Further, with our Blockwise Distillation method enhanced by Residual Feature Adaption and Teacher Attention Mechanism, we significantly increase the range of constructs that can be eliminated. For example, Layer Normalization can be mimicked by Batch Normalization. Finally, we arrive at a core set made of only five operators, comprising of only Convolution, Fully-Connected, Max-Pooling, Batch Normalization and Concatenation (both Channelwise and Spatial).

The rest of the paper is organized as follows: Sect.~2 discusses some work related to our own. We introduce the composition and construction of Arch-Net in Sect.~3. Finally, we give the results of experiments in Sect.~4. 

\section{Related work} \label{sec-relatedwork}

\textbf{Knowledge Distillation} Efforts have been made to transfer the learned knowledge from a large network to a smaller one by distillation since \cite{hinton2015distilling}. Following \cite{hinton2015distilling}, knowledge distillation methods focus on assimilating the logits distribution of the teacher and student network, by designing loss function \cite{hinton2015distilling}, distillation strategy \cite{zhang2018deep, shen2019meal} or regularization \cite{cho2019efficacy}. The intermediate features of the networks are then taken into consideration for better transferring the learned knowledge \cite{romero2014fitnets, zagoruyko2016paying, huang2017like}. There are also works delving into the relations between different features or data and forcing the student to mimic such relations \cite{yim2017gift, tung2019similarity, park2019relational}. Works that combine at least two of the above ideas have succeeded in tasks including classification \cite{shen2020meal}, object detection \cite{chen2017learning}, semantic segmentation \cite{liu2019structured, liu2019structureddesenprediction} and natural language processing \cite{sanh2019distilbert, wang2020minilm, jiao2019tinybert}. In our proposed method, the student network learns from the teacher's logits and intermediate features.

\textbf{Model Quantization} Networks of low bit width are computation, memory and energy efficient and therefore hardware friendly. Model Quantization methods compress the networks by mapping the real numbers to a set of discrete ones during or after training. Quantization-aware Training (QAT) methods quantize the networks by uniform \cite{zhou2016dorefa, cai2017deep, gong2019differentiable} or nonuniform \cite{polino2018model, zhu2016trained}, fixed \cite{courbariaux2016binarized, esser2016convolutional} or learned \cite{choi2018learning, choi2018pact, esser2019learned} quantizers during training. Although these methods enable the networks as low as 1 bit width \cite{courbariaux2016binarized, rastegari2016xnor} to have considerable classification or location performance, they require a similar training process as the training of floating-point networks, which is time and data consuming. It is worth mentioning that under some situations it is difficult to get all of the training data because of privacy concern. On the contrary, Post Training Quantization (PTQ) \cite{gholami2021survey} quantize the networks with a small calibration set of data sampled from the original training set \cite{banner2018aciq, zhao2019improving} or generated by special data generation methods \cite{nagel2019data, zhang2021diversifying}.
Recently, by introducing an optimizing or finetuning process \cite{hubara2020improving, nagel2020up, li2021brecq}, PTQ methods succeed in quantizing networks into as low as 4 bit width with a little loss of accuracy. However, when the bit width is lower, these methods seem powerless or need to turn to mixed precision \cite{dong2019hawq, dong2019hawqv2} for help. In addition, when the architectures of the networks are changed, PTQ methods become useless because no existing weights can be used for initialization or statistics. Compared to QAT and PTQ methods, our method enjoys the advantage of low bit width, less data, higher accuracy and architecture agnostic.

\textbf{Knowledge Distillation and Model Quantization} The combination of Knowledge Distillation and Model Quantization offers new ideas to model compression. With the help of distillation, the well learned knowledge of the floating-point networks can be transferred to the quantized ones. In \cite{mishra2017apprentice, polino2018model}, knowledge distillation is directly applied to training as low as 2W8A quantized networks. While in \cite{kim2019qkd}, a three phases method is adopted to get as low as 3W3A quantization of ResNet with little loss of accuracy. However, the data problem is still remaining because these methods rely on large amount of data with ground truth. What's more, these methods have not taken specific hardware constraints into consideration and therefore the quantized networks cannot be directly deployed in edge devices.

\begin{table}[]
\caption{The level of support of different chips or toolchains for basic units. `×' represents lack of support; numbers of `+' represents different levels of support; `?' means that the authors cannot ascertain the level of support. }
\label{chips-support-ability}
\footnotesize
\begin{tabular}{ccccccccc}
\hline
  & \begin{tabular}[c]{@{}c@{}}layer \\ norm\end{tabular} & embedding & \begin{tabular}[c]{@{}c@{}}depthwise \\ conv\end{tabular} & \begin{tabular}[c]{@{}c@{}}conv \\ 7×7\end{tabular} & \begin{tabular}[c]{@{}c@{}}channels\\ constraints\end{tabular} & \begin{tabular}[c]{@{}c@{}}\textbf{conv} \\ \textbf{3×3}\end{tabular} & \textbf{\begin{tabular}[c]{@{}c@{}}batch \\ norm\end{tabular}} & \textbf{FC} \\  \hline
\begin{tabular}[c]{@{}c@{}}Ethos-N Series\end{tabular} & ?                                                     & ?         & ?                                                         & ++      & ?                                                              & +++              & +++                                                            & +++             \\
Hi3559A                                                   & ×                                                     & ×         & ++                                                        & +       & +                                                              & +++              & +++                                                            & +++             \\
MLU270-S4                                                 & ×                                                     & ×         & ?                                                         & ++      & +                                                              & +++              & +++                                                            & +++             \\
TensorRT                                                  & +                                                     & +         & +                                                         & ++      & ++                                                             & +++              & +++                                                            & +++             \\
\begin{tabular}[c]{@{}c@{}}Movidius SDK\end{tabular}  & ?                                                     & ?         & +                                                         & +       & +                                                              & +++              & +++                                                            & +++             \\
SSC336Q                                                   & ×                                                     & ×         & ?                                                         & ++      & ++                                                             & +++              & +++                                                            & +++             \\
RK3568                                                    & ×                                                     & ×         & +                                                         & +       & +                                                              & +++              & +++                                                            & +++       \\    \hline 
\end{tabular}
\end{table}

\section{\methodname}
In this section, we firstly present the hardware constraints and the core operator set we build, followed by the transformation, quantization and initialization details. The Blockwise Model Distillation, including the Residual Feature Adaptation, the Teacher Attention Mechanism and the distillation algorithm are presented in turn.

\subsection{Hardware Constraints and Core Operator Set} \label{sec-transformation-set-and-model-transformation}
\begin{figure}
    \centering
    \includegraphics[width=1.0\textwidth]{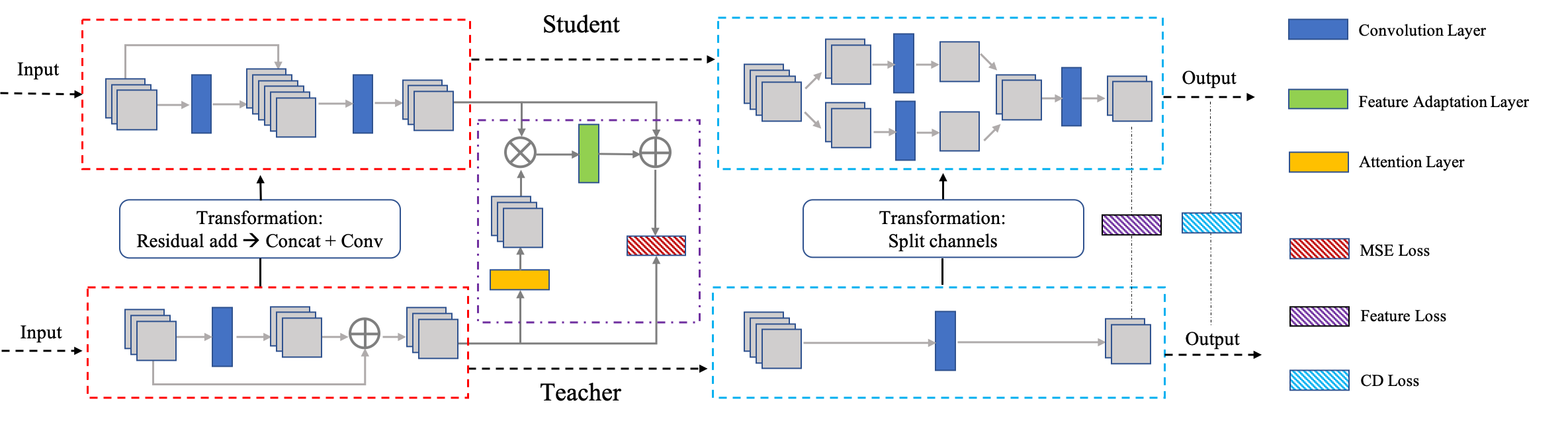}
    \caption{\methodname~pipeline for image classification}
    \label{fig-pipeline-for-image-classification}
\end{figure}

We have investigated ASICs/SDK/Toolchain from different companies and summarize our findings in Table \ref{chips-support-ability}. Overall the chips and their toolchains fall behind the development of DNN and can hardly support less common DNN constructs, like Layer Normalization and Embedding Layer which gain popularity due to increasing interests in Transformers. In addition, only convolutions of some kernel sizes are well supported \cite{Ding_2021_CVPR}. For example, we test the relative inference speed of common convolutions on SSC336Q chip from SigmaStar and found that 3×3 convolution is the most efficient (refer to Appendix \ref{sec-speed-test} for more details).
In order to bridge the gap between different DNN's and ASIC chips, we build the following core operator set, which contains only the well supported operators. 
\begin{itemize}
    \item 3×3 Convolution of stride 1 or 2: we use it to replace all of the other convolutions.
    \item 2x2 Max-pooling of stride 2: we use it replace all of the other max-pooling layers.
    \item Batch Normalization \cite{ioffe2015batch}: we use it to replace Layer Normalization.
    \item Fully-Connected layer: we use it to replace the Embedding layer.
    \item We assume the limitation for channel number is 512 and split the channel when the number of input/output channels is larger than 512 (Figure \ref{fig-pipeline-for-image-classification}).
    \item Concatenation + 3×3 convolution: we use it to replace the residual addition (Figure \ref{fig-pipeline-for-image-classification}).
\end{itemize}

\begin{figure}
    \centering
    \includegraphics[width=1.0\textwidth]{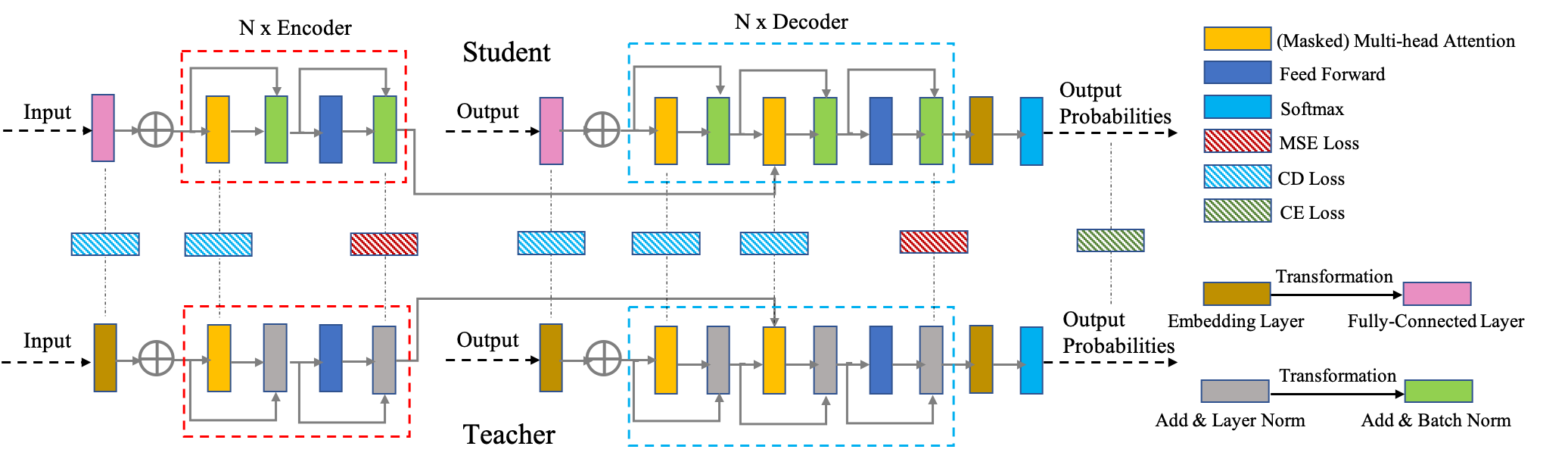}
    \caption{\methodname~pipeline for machine translation}
    \label{fig-pipeline-for-machine-translation}
\end{figure}

Based on this core operator set, we can transform the floating-point networks into simple ones, namely the Arch-Net. For example, we use three 3×3 convolutions to replace a 7×7 convolution. For the Layer Normalization (LN) in Transformer \cite{vaswani2017attention}, we replace the it with Batch Normalization (BN) (Figure \ref{fig-pipeline-for-machine-translation}). To be specific, assuming the shape of the feature before the LN is $[B, W, D]$ (where $B$ is the batch size, $W$ is the number of words, and $D$ is hidden dimension), we permute it to be $[B, D, W, 1]$ and feed it to the BN to get a output of shape $[B, D, W, 1]$. The output is then quantized and permute back to the shape of $[B, W, D]$. For the embedding layer, it is intuitive to use a fully-connected layer to replace it. As a result, the input data of shape $[B, D]$ should be converted to one-hot form $[B, D, S]$, where $S$ is the vocabulary size. The other transformations details can be found in Appendix \ref{appendix-transformation-details}.

We choose DoReFa-Net \cite{zhou2016dorefa} as the quantization method for classification tasks and LSQ \cite{esser2019learned} for machine translation tasks, to show the robustness of our method to different quantization methods. 
Before training, the Arch-Net will be initialized with the weights from the floating-point teacher. For the 3×3 convolutions that replace the 3×3 convolutions, we directly copy the floating-point weights for the student. For the 3×3 convolutions that replace the 1×1 convolutions, the floating-point weights are padded to 3×3 with zeros. The other operators that do not have corresponding weights are initialized randomly.

\subsection{Blockwise Model Distillation}
\textbf{Algorithm} Different from most of the knowledge distillation framework \cite{hinton2015distilling, shen2019meal, tung2019similarity, sanh2019distilbert, wang2020minilm}, we distill the network in a progressive and block-wise manner, which is architecture agnostic. Inspired by the Block Coordinate Descent algorithm \cite{tseng2001convergence}, we divide the network into blocks, and optimize the student's blocks stage by stage, by minimizing the outputs of the student's and teacher's blocks. The difference is that when optimize a block, the previous blocks are also optimized at the same time. We find that other frameworks does not work well when the architecture of the student network is changed and the bit width is low. We argue that it is because the network has not been well initialized although some of the floating-point weights from the teacher can be directly copied for the student. In addition, when the architecture is changed, no existing weights can be used for initialization. We therefore adopt this block-wise and progressive distillation framework for better initialization. The algorithm is shown in Algorithm \ref{alg:layerwiseframework}.

\textbf{ResNet and MobileNet} Take ResNet18 as example. For the first 3 convolutional layers and the 2x2 max-pooling layer of the student, we put them together as $\mathit{block 1}$ and force it to learn from the output feature of the $\mathit{block 1}$ of the teacher, which consists of the first 7×7 convolutional layer and a 3×3 max-pooling layer. The above distillation procedure is referred as $\mathit{stage 1}$. For each of the next stages but the last one, the block contains Convolution + BN + ReLU. For the last stage, we put the average pooling layer and the fully-connected layer together as one block. Starting from $\mathit{stage 2}$, the weights of the student is loaded from the previous stage. The block-wise Distillation for MobileNet is similar to that for ResNet, except that there is no 7×7 convolution and max-pooling layer. For the last stage, we use the Cosine Distance
as the objective function. For other stages, we use the Mean Square Error (MSE) between the two feature maps. As we want to focus on the distillation of the current block, we reduce the importance of the previous blocks by adding a weighting coefficient $\gamma \in (0, 1)$. Assuming that there are $n$ stages, at stage $m$, the objective function is given as Equation (\ref{totalloss}).


\begin{gather}
    \mathit{loss} = \sum_{i=1}^{m} \gamma^{m-i}\mathit{loss}_{i} \label{totalloss} \\
    \text{where}
    \left\{
        \begin{array}{lr}
             \mathit{loss}_{i} = \text{MSE}(F_S^i, F_T^i),  &i \in [0,n-1]  \\
             \mathit{loss}_{i} = \text{D}_{\cos}(L_S^i, L_T^i), & i = n
        \end{array}
    \right. \notag
\end{gather}
Where $F_S^i$ and $F_T^i$ are the output featuremaps, $L_S^i$ and $L_T^i$ are the output logits before softmax of the $i^{\text{th}}$ stage of the student and the teacher network, separately. It is worth mentioning that as we directly use the output logits of the teacher, no ground truth is needed.

\textbf{Transformer} Each encoder or decoder is one distillation block, and the last block contains the final projection layer. For each block except for the last block, we force the output of each BN in the student to assimilate that of each LN in the teacher and MSE is adopted as objective function. For the last block, we take the teacher's predictions as hard label ($y_{\mathrm{hard}}$) so that the Cross Entropy (CE) loss can be used. Besides, the attention maps and the outputs of the embedding layer are also used for distillation, and the cosine distance is used here. The final objective function can be formulated as follows:
\begin{gather}
    \mathit{loss} = \sum_{i=1}^{m} \gamma^{m-i}\mathit{loss}_{i} \label{totalloss_transformer} \\
    \text{where}
    \left\{
        \begin{array}{lr}
                 \mathit{loss}_{i} = \sum_{j} \text{MSE}(N_S^{ij}, N_T^{ij}) + \sum_{k} \text{ D}_{\cos}(A_S^{ik}, A_T^{ik}) +
                 \text{D}_{\cos}(E_S, E_T),  &i \in [0,n-1]  \\
             \mathit{loss}_{i} = \text{CE}(L_S^i, y_{\mathrm{hard}}), & i = n
        \end{array}
    \right. \notag
\end{gather}
Where $N_S$ and $N_T$ represent the outputs of the normalization layer of the student and the teacher, $A_S$ and $A_T$ are the attention maps, $E_S$ and $E_T$ are the outputs of the embedding layer, $i$ represents the $i^{\text{th}}$ distillation stage. As there are more than one LN and attention map in each block, we use $j$ and $k$ to represent the number. $L_S$ represents the final output of the student.

For image classification tasks, as there is already a well trained teacher, and the student inherits most of the teacher's weights, we claim that a modest number of images is enough. For the whole distillation procedure, we randomly sample 30k images without ground truth from ImageNet \cite{russakovsky2015imagenet} training set. Besides, as the stages but the last stage are for better initialization, they require a smaller number of training epochs. In addition, for each epoch, we adopt a training strategy of randomly sampling a small number of images from the 30k images. Further details can be found in Section \ref{sec-experiments}. Under this setting, the need of data is reduced and the training time is greatly shortened.

\subsection{Residual Feature Adaptation and Teacher Attention Mechanism}
The difficulty of the knowledge distillation between a floating-point teacher and a quantized student results from the fact that the output value range of these two networks is quite different. The quantization method we use clips the output value of the student into $[0,1]$, while the output value of the floating-point operator is much larger. This gap brings the difficulty for the student in mimicking the output distribution of the teacher. In works \cite{romero2014fitnets, chen2017learning} of distillation between two floating-point networks, several convolutional or fully-connected layers are adopted to map the features of the student to that of the teacher, including the mapping of sizes and latent representations. In \methodname, similar operation is adopted to map not only the latent representations, but also the output value range. We apply a Residual Feature Adaptation (RFA) block of 3 continuous floating-point convolutional (for ResNet/MobileNet) or fully-connected (for Transformer) layers plus a residual addition, without any batch normalization or non linearity, as shown in Figure  \ref{fig-pipeline-for-image-classification}. We find the residual addition important because the outputs of the feature adaptation blocks is not directly used for inference, and the auxiliary block-wise losses suffer from the vanishing gradient problem without it. By adding the student's features to the output of the feature adaptation block, not only the value range, but also the latent representations of the teacher's features can be better mimicked by the student. For ResNet and MobileNet, we add the Residual Feature Adaptation at the output of each block except for the first and the last block. For Transformer, we add it after each BN. The ablation studies in Section \ref{sec:ablation-study} show the importance of the residual additions. Floating-point operators are adopted here because they are able to better map the value ranges from fixed-point numbers to real numbers.

In addition to the Residual Feature Adaptation Block that helps the student to mimic the feature distribution of the teacher, we think it is also crucial for the student to learn the importance of each channel. It is intuitive that we can extract a series of weighting coefficients of the channel and feed them to the student. Following the attention mechanism in SENet \cite{hu2018squeeze}, we feed the teacher's features to a block of 2 fully-connected layers to get a sequence of channel weighting coefficients and then multiply these coefficients and the student's feature. The output of the multiplication is then fed to the Feature Adaptation layers, as shown in Figure \ref{fig-pipeline-for-machine-translation}. As the weighting coefficients come from the teacher, we name it Teacher Attention Mechanism (TAM). 

The RFA and the TAM bridge the gap between the quantized student and the floating-point teacher, enabling the student to learn from the distribution of the teacher's features and the channel-wise relations. In addition, this two methods are applied only during the distillation, bringing no extra computational cost during inference time.
\begin{algorithm}[htb] 
\caption{ Blockwise Model Distillation.} 
\label{alg:layerwiseframework} 
\begin{algorithmic}[1]
\REQUIRE ~~\\
    Floating-point teacher network $T_f$;\\
    Arch-Net $S_q$;\\
    Number of images for the whole distillation procedure $P_{total}$;\\
    Number of images sampled per epoch $P_{epoch}$;\\
    Number of stages $N$;\\
    Number of epochs for the last stage $E_{last}$;\\
    Number of epochs for the stages except for the last stage $E_{middle}$;\\
\ENSURE ~~\\
	Well trained Arch-Net $S_q$;
	\STATE Initialize $S_q$ with $T_f$, randomly sample $P_{total}$ images from the training set
	\FOR{$\mathit{Stage}=1, \cdots, N$}
	\FOR{$\mathit{Epoch}=1, \cdots, E_{last}$ if $\mathit{Stage} == N$ else $E_{middle}$ }
	\STATE Randomly sample $P_{epoch}$
	\STATE Update $S_q$ by minimizing objective function (\ref{totalloss}) and (\ref{totalloss_transformer})
	\ENDFOR
	\ENDFOR
\RETURN $S_q$;
\end{algorithmic}
\end{algorithm}

\section{Experiments} \label{sec-experiments}
In this section, we perform experiments on ImageNet \cite{russakovsky2015imagenet} and Multi30k \cite{elliott2016multi30k} to evaluate the performance of our proposed method. Networks used in the experiments include ResNet18/34/50 \cite{he2016deep}, MobileNet V1 \cite{howard2017mobilenets}, MobileNet V2 \cite{sandler2018mobilenetv2} for image classification task, and Transformer \cite{vaswani2017attention} for machine translation task. 
We show that the proposed \methodname~is able to transform the previous floating-point networks into as low as 2W4A hardware-friendly Arch-Net with as low as 0.9\% loss of accuracy for ImageNet classification tasks and no loss of BLEU score for machine translation tasks. Comprehensive ablation studies are also conducted to further investigate the proposed method.

\subsection{Experiments Setup}
We perform out experiments on NVIDIA Tesla V100 GPU in PyTorch. For the image classification task, we randomly sample 30k images from the original training set as our training sets. At each training epoch, we randomly sample 8192 images from our training set. For the machine translation task, we use the whole Multi30k dataset as our training set and we use all of the data at each training epoch. Multi30k is a small dataset while the vocabulary is relatively large, it will be helpful to use the whole training set.

The optimizer we use for the image classification task is Adam \cite{Diederik2015Adam} with a learning rate of 1e-3 and the learning rate scheduler is Cosine Annealing with Warm Restart \cite{loshchilov2016sgdr}, that for the machine translation task is AdamW \cite{loshchilov2018decoupled} with the same learning rate and scheduler. For ResNet18/34/50, the number of epochs for the first stage is 500 (as the first stage contains three 3×3 convolutions that replace the 7×7 convolution, which is not initialized and needs more epochs), that for the last stage is 5110 (5110 is the convergence point of the learning rate scheduler we use) and that for the other stages is 60. For MobileNetV1/V2, the number of epochs for the last stage is 5110, that for the other stages is 60. For Transformer, the number of epochs for the middle stages is 6, and that for the last stage is 50. The source of the teacher networks can be found in Appendix \ref{source-of-the-teacher-networks}.


\subsection{Results on ImageNet}  \label{sec:results-on-imagenet}
We evaluate our proposed method on ImageNet dataset and the results are shown in Table \ref{results-on-imagenet-ptp}. We firstly compare our method with the DoReFa-Net trained with the whole dataset, and trained with 30k images randomly sample from the whole dataset because we use DoReFa-Net as our quantization method. It is interesting that the results of QAT is worse than that of \methodname, even though the former method use the whole ImageNet dataset as training set while the latter use only 30k images without ground truth. We conclude that it reflects the positive influence of the knowledge distillation, where the well learned knowledge is transferred from the floating-point teacher to the quantized student. To be specific, for the reason of hardware-friendly, the structure of the student is greatly restricted. Therefore, without the diversity of the convolutional layers with different kernel sizes and without the residual addition which is helpful for training, it is difficult to train (QAT) a good student of low bit width (2W4A). On the contrary, knowledge distillation is able to empower this student, even with fewer data and even the structure of the student is greatly changed. For all of the students of these 5 floating-point networks, the loss of accuracy is less than 1.6\%. For ResNet34, the loss of accuracy is only 0.9\%. We attributes this gain to the Residual Feature Adaptation Block, Teacher Attention Mechanism and the Blockwise Distillation. Further analysis is provided in Section \ref{sec:ablation-study}.
 
\begin{table}[H]
    \scriptsize
    \caption{Imagenet performance of Arch-Net.}
    \label{results-on-imagenet-ptp}
    \centering
    \begin{tabular}{cccccccccccc}
    \toprule
    \multirow{2}{*}{Method} & \multirow{2}*{Bit width}     & \multicolumn{2}{c}{Resnet18} & \multicolumn{2}{c}{Resnet34}& \multicolumn{2}{c}{Resnet50}& \multicolumn{2}{c}{Mobilenet V1}& \multicolumn{2}{c}{Mobilenet V2} \\
    \cmidrule(r){3-4}  \cmidrule(r){5-6} \cmidrule(r){7-8}  \cmidrule(r){9-10} \cmidrule(r){11-12}
    &&Top1 & Top5 & Top1 & Top5 & Top1 & Top5 & Top1 & Top5 & Top1 & Top5 \\
    \midrule
    Teacher    & 32W32A      & 69.76     & 89.08 &73.30&91.42&76.13&92.86&68.79&88.68&71.88&90.29 \\
    Dorefanet (30k) & \multirow{3}{*}{2W4A}& 29.93 & 54.65 &12.00&27.95&8.20&20.62&21.28&43.20&6.22&16.38\\
    Dorefanet (whole) & & 67.19 & 87.68 &67.46&87.78&66.82&87.59&64.36&86.06&59.89&82.77\\
    Ours (30k) & & \textbf{68.77} & \textbf{88.66} &\textbf{72.40}&\textbf{91.01}&\textbf{74.56}&\textbf{92.39}&\textbf{67.29}&\textbf{88.07}&\textbf{69.09}&\textbf{89.13}\\
    \bottomrule
    \end{tabular}
\end{table}

In order to compare our method with other QAT and PTQ methods, we apply no previous transformations to the teachers but quantize them to 2W4A students based on DoReFa-Net, and then use the Blockwise Model Distillation to train the students. The architectures of the quantized students are therefore the same as that of the teachers, so that they are comparable to other QAT and PTQ methods. Here, we compare our method with QAT methods includes DoReFa-Net, LSQ \cite{esser2019learned}, DSQ \cite{gong2019differentiable}, QKD \cite{kim2019qkd} and PTQ method includes BRECQ \cite{li2021brecq}, PWLQ \cite{fang2020post} and ZeroQ \cite{cai2017deep}. As shown in Table \ref{results-on-imagenet-on-qat-ptq}, for ResNet, \methodname~outperforms these methods by a big advantage. While for MobileNet, LSQ and DSQ have better performance. We argue that it is because they use the whole training set of ImageNet while we use only 30k. We show in Table \ref{results-on-imagenet-on-qat-ptq} that when they use 30k images, the results are much worse. Besides, the quantization method we use, the DoReFa-Net is quite old that it also brings some loss of accuracy. On the whole, compared with the QAT methods, \methodname~requires less data. Compared with other PTQ methods, \methodname~enjoys a higher accuracy. Results of higher bit width are shown in Appendix \ref{sec-higher-bit-width}.
\begin{table}
    \scriptsize
    \caption{Imagenet performance of different QAT and PTQ methods without transformations.}
    \label{results-on-imagenet-on-qat-ptq}
    \centering
    \begin{tabular}{cccccccccccc}
    \toprule
    \multirow{2}{*}{Bit Width} & \multirow{2}*{Method}     & \multicolumn{2}{c}{ResNet18} & \multicolumn{2}{c}{ResNet34}& \multicolumn{2}{c}{ResNet50}& \multicolumn{2}{c}{MobileNet V1}& \multicolumn{2}{c}{MobileNet V2} \\
    \cmidrule(r){3-4}  \cmidrule(r){5-6} \cmidrule(r){7-8}  \cmidrule(r){9-10} \cmidrule(r){11-12}
    &&Top1 & Top5 & Top1 & Top5 & Top1 & Top5 & Top1 & Top5 & Top1 & Top5 \\
    \midrule
    32W32A  & Teacher &  69.76     & 89.08 &73.30&91.42&76.13&92.86&68.79&88.68&71.88&90.29\\
    \cmidrule(r){2-12}
    \multicolumn{12}{c}{\qquad \qquad \quad PTQ} \\
    \cmidrule(r){2-12}
     \multirow{13}{*}{2W4A}   & BRECQ   & 64.42&86.22 &--&--&69.67&89.47&--&--&18.17&38.66\\
    & PWLQ & 19.31 & 38.95 &27.01&48.75&27.90&43.28&--&--&--&--\\
    & ZeroQ & 0.13 & 0.51 &0.09&0.48&0.10&0.51&--&--&0.09&0.52\\
    \cmidrule(r){2-12}
    \multicolumn{12}{c}{\qquad \qquad \qquad QAT - whole training set} \\
    \cmidrule(r){2-12}
    & DoReFa-Net  & 60.46 & 83.25 &65.93&86.64&66.33&87.35&48.24&73.14&44.87&69.93\\
    & LSQ  &63.69 & 84.75 &66.98&87.11&70.23&89.54&66.25&86.58&61.83&83.57\\
    & DSQ  & 58.18 & 81.40 &62.61&84.42&68.23&88.74&--&--&62.72&84.90\\
    & QKD  &64.48 & 85.68 &68.76 &88.08 &69.39&88.52&--&--&41.21&66.08\\
    \cmidrule(r){2-12}
    \multicolumn{12}{c}{\qquad \qquad \qquad QAT - 30k images} \\
    \cmidrule(r){2-12}
    & LSQ  &1.86 &2.29 &1.87&2.30&1.90&2.29&1.70&2.24&1.53&2.14\\
    & DSQ  &10.61&25.48 &12.75 &29.03&12.67&29.08&--&--&13.71&31.11\\
    & QKD  &23.90& 56.37&23.49&56.21&31.04&65.78&--&--&1.83&3.80\\
    & Ours  & \textbf{67.30} &\textbf{87.73} &\textbf{71.58} &\textbf{90.51}&\textbf{74.59}&\textbf{92.29}&\textbf{59.66}&\textbf{83.31}&\textbf{57.63}&\textbf{82.00}\\
    \bottomrule
    \end{tabular}
\end{table}

\subsection{Results on Multi30k}
The results are shown in Table \ref{results-on-multi30k}. Here in the quantized Transformer, we do not replace the residual additions for the reason that there are fully-connected layers in Transformer, if we use concatenating layer + fully-connected layer to replace the residual addition, the computational overhead would be large. The results show that with LN and embedding layers replaced, quantized \methodname~performs even better than the float teacher. We ascribe this to the initialization of the Blockwise Model Distillation and the alignment of the featuremaps and the output logits.

\begin{table}[H]
    \caption{Results on Multi30k}
    \label{results-on-multi30k}
    \centering
    \setlength{\abovecaptionskip}{0.cm}
    \begin{tabular}{cccccc}
    \toprule
    Networks    & Bit Width  & Task & BLEU & Task & BLEU \\
    \midrule
    \multirow{4}*{Transformer}   & 32W32A  &\multirow{4}*{DE-EN}   & 30.32 &\multirow{4}*{EN-DE} & 32.44 \\
    & 8W8A &&34.05&&36.44 \\
    & 4W4A &&34.34&&34.35\\
    & 2W4A &&32.50&&33.75\\
    \bottomrule
    \end{tabular}
\end{table}

\subsection{Ablation Study}  \label{sec:ablation-study}

\begin{figure}[H]
\centering
\subfigure[Top1 Accuracy with/without TAM]{
\label{fig:comparison-teacher-attention}
\includegraphics[height=0.32\textwidth]{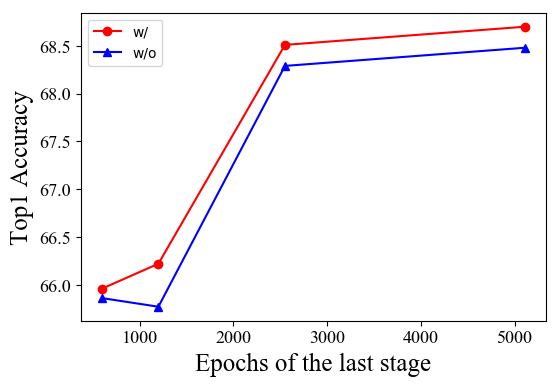}}
\subfigure[Top1 Accuracy of different middle stage epochs]{
\label{fig:top1-defferent-epochs-middle-stages}
\includegraphics[height=0.32\textwidth]{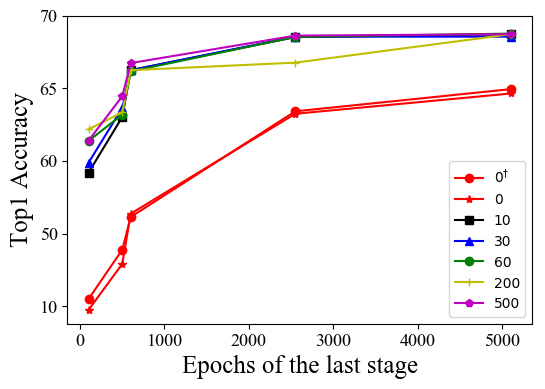}}
\caption{Comparison of Top1 Accuracy}
\label{Fig-comparison-of-top1-accuracy}
\end{figure}

\textbf{Importance of the middle epochs in Blockwise Model Distillation (BMD)} Is BMD necessary? Will it help to use more epochs for middle stages or harm to use less? As we mentioned above, the purpose of middle stage distillation is a better initialization. Here, we test different numbers of epochs for each middle stage (all stages but the first and the last), as shown in Figure \ref{fig:top1-defferent-epochs-middle-stages}. At the situation of curve $\text{0}^\dag$ in the figure, the BMD turns to be an one-shot distillation, and the result turns out to be inferior. On the other hand, the result is much better when the numbers of epochs is non-zero, even if it's as few as 10. It confirms our previous argument that the middle stages of the BMD are for better initialization. While when the initialization is done, more epochs for the middle stages will not bring more gain in accuracy.

\begin{figure}
    \centering
    \includegraphics[width=1.0\textwidth]{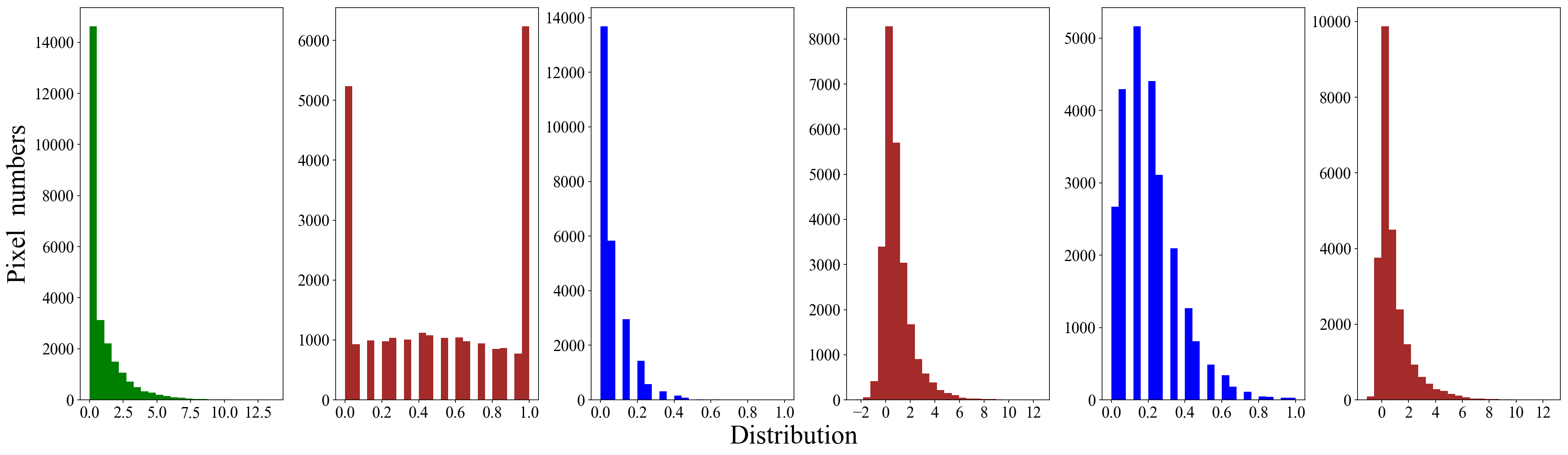}
    \caption{Comparison of the output distributions. Use 25th block of ResNet18 as an example. Left to right: teacher, student w/o feature adaptation, student before feature adaptation, student feature Adaptation, student before residual feature adaptation, student after residual feature adaptation.}
    \label{fig-compare-fa}
\end{figure}

\textbf{Importance of Residual Feature Adaptation} We directly remove the residual addition from the Residual Feature Adaptation block, and we also use only the BMD without any feature adaptations. The results are shown in Table \ref{ablation-residual-feature-adaptation}. The residual addition helps the mapping of the teacher and the student's featuremaps, especially when the data is not sufficient (trained with 1024 images per epoch). In addition, we count the distribution of the values of the featuremaps of the student and teacher in Figure \ref{fig-compare-fa}, showing that the Residual Feature Adaptation is better at mapping the value range of the student's featuremaps to that of the teacher's.
\begin{table}[H]
    \scriptsize
    \caption{Comparison of the feature adaptation with/without residual addition. Take ResNet18 of 2W4A as example.}
    \label{ablation-residual-feature-adaptation}
    \centering
    \begin{tabular}{ccccccc}
    \toprule
    Feature Adaptation & Images Per Epoch & Top1 & Top5  & Images Per Epoch & Top1 & Top5 \\
    \midrule
    w/ residual addition & \multirow{3}*{1024} & 67.60 & 88.32 & \multirow{3}*{8192} & 68.77 & 88.66 \\
                              w/o residual addition  && 65.41 & 87.47 & & 68.52 & 87.44 \\
                              None               && 62.43 & 85.96 & & 64.09 & 86.49 \\
    \bottomrule
    \end{tabular}
\end{table}

\textbf{Importance of Teacher Attention Mechanism (TAM)} We find that the Teacher Attention Mechanism brings a slight increase in accuracy, 
and it is also helpful in helping the convergence of the student, as shown in Figure \ref{fig:comparison-teacher-attention}, reflecting the importance of the learning of channel-wise relation.

\section{Conclusion}
In this paper, we propose \methodname~to bridge the gap between Computer Architecture of ASIC chips and Neural Network Model Architectures, by transforming existing floating-point DNNs into hardware-friendly quantized \methodname, with bit-width as low as 2W4A. The structure of \methodname~is constructed from the core operator set that consists of only five operators: 3×3 Convolutions, Batch Normalization, Concatenation, 2x2 Max-pooling, and Fully-Connected layers, which are so common, that they are often the least restricted and most efficient operators in the ASIC chips we tested. Labeled data is not required for the conversion to \methodname~as we employ Blockwise Model Distillation on feature maps. Extensive experiments on image classification and machine translation tasks confirm that \methodname~is both effective and data-efficient. For example, we are able to transform floating-point networks into 2W4A quantized ones with the loss of accuracy as low as 0.9\% on ImageNet Classification tasks, and no loss of BLEU score on Multi30k.

\printbibliography

\clearpage
\appendix

\section{Appendix} \label{sec-appendix}

\subsection{Main Constraints of Popular ASIC Chips, SDK and Toolchains for Quantized Networks} \label{sec-appendix-constraints}
We survey some of the popular ASIC chips, SDK and Toolchains for quantized networks and summarize in Table \ref{table-chips-constraints}. As the public information is difficult to get, we summarize from what we can reach. This table is complementary to Table \ref{chips-support-ability}.

\begin{table}[H]
  \caption{Main constraints of popular ASIC/SDK/ToolChain for quantized networks.}
  \label{table-chips-constraints}
  \centering
  \begin{tabular}{ccc}
    \toprule
    Company     & Type      & Main Constraints \\
    \midrule
    ARM         & Ethos-N Series  & Only support Int16 and Int 8    \\
    \specialrule{0em}{1pt}{1pt}
    \cline{3-3}
    \specialrule{0em}{1pt}{1pt}
    Hisilicon   & Hi3559A         & \makecell[l]{1. As low as Int8 and Uint8 are supported \\
                                    2. The Number of channels in convolutional layers \\ \quad is suggested to be the multiples of 32  \\
                                    3. Batch Normalization is suggested to be used as \\ \quad normalization layer while Layer Normalization is not \\
                                    4. Early Networks (VGG, Alexnet, etc.) are not \\ \quad suggested 
                                    to be used \\
                                    5. Too many pooling layers among convolutional layers \\ \quad will harm the networks' performance}    \\
    \specialrule{0em}{1pt}{1pt}
    \cline{3-3}
    \specialrule{0em}{1pt}{1pt}
    Cambricon   & MLU270-S4       & As low as Int4 is supported  \\
    \specialrule{0em}{1pt}{1pt}
    \cline{3-3}
    \specialrule{0em}{1pt}{1pt}
    Nvidia      & TensorRT        & As low as Int8 is supported \\
    \specialrule{0em}{1pt}{1pt}
    \cline{3-3}
    \specialrule{0em}{1pt}{1pt}
    Intel       & \makecell[c]{Movidius \\ Neural Compute SDK}  & \makecell[l]{1. Group number of group                                      convolutions needs to be \\ \quad less than 1024 \\
                                                 2. 5x5 convolution of stride 2 is not supported}  \\
    \specialrule{0em}{1pt}{1pt}
    \cline{3-3}
    \specialrule{0em}{1pt}{1pt}
    SigmaStar &   SSC336Q   & \makecell[l]{1. Except for the first convolutional layer, the number \\
                                    \quad of input/output channels should be less than 2048 \\
                                    2. Depthwise convolutions: only kernel size of 3x3 \\ \quad is supported} \\
    \specialrule{0em}{1pt}{1pt}
    \cline{3-3}
    \specialrule{0em}{1pt}{1pt}
    Rockchip &   RK3568 & \makecell[l]{1. 3x3 Convolution is suggested to be used\\
                                    2. Maxpooling: only 2x2 of stride 2 and 3x3 of stride 2 \\
                                    \quad are supported} \\
    \bottomrule
  \end{tabular}
\end{table}

\subsection{Speed Test on ASIC Chips} \label{sec-speed-test}
\subsubsection{SSC336Q ASIC Chip}
We use the SSC336Q chip from SigmaStar to test the speed of basic convolutions. Backbones with only one of certain basic convolution are set up. In order to run on the chip successfully, a low-calculation tail is added after backbones.  We also use two 3x3 convolutions to replace the 5x5 convolution and three 3x3 convolutions to replace the 7x7 convolution, and keep the FLOPs similar as the original network. The running time on SSC336Q are shown in Table \ref{inference-speed-sigmastar}. It is clear that the inference speed of two 3x3 convolutions is faster than that of a single 5x5, and that of three 3x3 convolutions is faster than that of a single 7x7, which provides proof for the transformation we make in this paper.
\begin{table}[H]
    \caption{Inference speed of different convolutions on Sigmastar SSC336Q}
    \label{inference-speed-sigmastar}
    \centering
    \begin{tabular}{ccccccc}
    \toprule
    Basic convolutions & 1x1 & 3x3 & 5x5 & two 3x3  & 7x7 & three 3x3\\
    \cmidrule(r){2-3} \cmidrule(r){4-5}  \cmidrule(r){6-7}
    Speed (ms)     & 0.691 & 0.906 & 2.370  & 1.670 & 3.889 &  2.538  \\
    \bottomrule
    \end{tabular}
\end{table}

\subsubsection{Rockchip RK3568 ASIC Chip}
We also carry out experiments on Rockchip RK3568 ASIC chip. Firstly, we choose VGG19 to verify the replacement of Max-pooling layer. We direcly replace the 2x2 Max-pooling layers in VGG19 with 3x3 Max-pooling layers test the inference speed. As shown in Table \ref{results-of-vgg19-on-rk}, VGG19 with 2x2 Max-pooling layers performs slightly better than that with 3x3 Max-pooling layers in inference speed and consumes less time in each Max-pooling Layer except for the last one.

\begin{table}[H]
    \scriptsize
    \caption{Inference speed of VGG19 on Rockchip RK3568}
    \label{results-of-vgg19-on-rk}
    \centering
    \begin{tabular}{cccccccc}
    \toprule
    \multirow{2}{*}{Max-pooling type} & \multirow{2}*{Bit width} & \multirow{2}*{Inference Speed (fps)} & \multicolumn{5}{c}{Time usage (ms) of each Max-pooling layer} \\
    \cmidrule(r){4-8}
    &&& Layer1 & Layer2 & Layer3 & Layer4 & Layer5 \\
    \midrule
    2x2 & \multirow{2}{*}{8W8A} & 13.82 &3.415&1.737&0.922&0.557&0.254  \\
    3x3 & & 13.45 &4.569&2.420&1.231&0.658&0.238 \\
    \bottomrule
    \end{tabular}
\end{table}

\subsubsection{Inference Performance of Arch-Net on Different ASICs}
We apply our transformations to Resnet18 and Mobilenet V2 to get the corresponding Arch-Net. We calculate the FLOPs (the number of Multiply-Adds) and compare the FLOPs per millisecond between the original networks and the Arch-Nets, under the situation of 8W8A, on different ASICs, including Sigmastar SSC336Q, Rockchip RK3568 and Nvidia Tesla T4. As shown in Table \ref{inference-performance-of-archnet-in-chips}, the FLOPs per millisecond of Arch-Net is much better than the original networks \footnote{For the Arch-Net of Mobilenet V2 on SSC336Q, we do not expand the number of channel in the Inverted Residual Bottleneck as the original Mobilenet v2 does (expand 6 times) because we fail to run the expanded one on SSC336Q (We guess it is because there is a upper limit for the FLOPs or Parameters).}.

\begin{table}[H]
    \scriptsize
    \caption{Inference performance of Arch-Net on different ASICs}
    \label{inference-performance-of-archnet-in-chips}
    \centering
    \begin{tabular}{ccccccccc}
    \toprule
    &\multirow{2}*{Networks} & \multirow{2}*{FLOPs (G)} & \multicolumn{2}{c}{SSC336Q} & \multicolumn{2}{c}{RK3568} & \multicolumn{2}{c}{Tesla T4} \\
    \cmidrule(r){4-5} \cmidrule(r){6-7} \cmidrule(r){8-9}
    & & & Speed (ms) & FLOPs/ms & Speed (ms) & FLOPs/ms & Speed (ms) & FLOPs/ms  \\
    \midrule
    \multirow{2}*{Resnet18} & Original & 1.82 & 15.68 &0.116&18.52&0.098&0.81&2.236 \\
    & Arch-Net & 5.89 & 38.11 &\textbf{0.154}&41.67&\textbf{0.141}&1.52&\textbf{3.836} \\
        \cmidrule(r){2-9}
    \multirow{2}*{MobilenetV2} & Original & 0.31 &8.23&0.037&19.23&0.016&0.78&0.396 \\
    & Arch-Net & 2.41 &9.70&\textbf{0.070}&55.56&\textbf{0.043}&1.91&\textbf{4.140} \\
    \bottomrule
    \end{tabular}
    
\end{table}


\subsection{Transformations Details} \label{appendix-transformation-details}
For Resnet18/34/50, the 7x7 convolution of stride 2 in the first layer is replaced by three 3x3 convolutions, of which the 3rd convolution is of stride 2 and the others are of stride 1. All of the other 1x1 and 3x3 convolutions are replaced by 3x3 convolutions. Besides, the 3x3 max-pooling of stride 2 is replaced by a 2x2 max-pooling of stride 2. For each residual block, the residual add is replaced by a concatenating layer and a 3x3 convolution of stride 1. When the number of the channels of a featuremap (which is the input of a convolutional layer) is larger than 512, we split it into several featuremaps with channels smaller than or equal to 512, feed them to parallel branches, each of which consists of one 3x3 convolution, and then concatenate the output featuremaps to get a new featuremap that has the same number of channels as that before the split (Figure \ref{fig-pipeline-for-image-classification}).

For Mobilenet V1 and Mobilenet V2, similar operations are performed. We replace the depthwise and pointwise convolutions with 3x3 convolutions. Residual adds in Mobilenet V2 are replaced by concatenating layers and 3x3 convolutions. Convolutions with number of channels that is large than 512 are split as what we do to Resnet.

For Transformer, except for the Layer Normalization, we use a fully connected layer to replace the embedding layer.

\subsection{Source of the Teacher Networks} \label{source-of-the-teacher-networks}
As for the teacher networks, we use the well trained ResNet18/34/50 and MobileNet V2 from Torchvision \footnote{https://s3.amazonws.com/pytorch/models}, and we train MobileNet V1 and Transformer ourselves based on codebase \footnote{https://github.com/jadore801120/attention-is-all-you-need-pytorch} and codebase \footnote{https://github.com/wjc852456/pytorch-mobilenet-v1}.

\subsection{Results of Higher Bit Width on ImageNet} \label{sec-higher-bit-width}
We have already proved the effectiveness of our method at 2W4A, here we show the results of our experiments of high bit width. It is intuitive that as the bit width becomes bigger, the transformed networks perform better. 

\begin{table}[H]
    \scriptsize
    \caption{Imagenet performance of higher bit width, with transformations.}
    \label{results-on-imagenet-high-bit}
    \centering
    \begin{tabular}{cccccccccccc}
    \toprule
    \multirow{2}{*}{Method} & \multirow{2}*{Bit width}     & \multicolumn{2}{c}{Resnet18} & \multicolumn{2}{c}{Resnet34}& \multicolumn{2}{c}{Resnet50}& \multicolumn{2}{c}{Mobilenet V1}& \multicolumn{2}{c}{Mobilenet V2} \\
    \cmidrule(r){3-4}  \cmidrule(r){5-6} \cmidrule(r){7-8}  \cmidrule(r){9-10} \cmidrule(r){11-12}
    &&Top1 & Top5 & Top1 & Top5 & Top1 & Top5 & Top1 & Top5 & Top1 & Top5 \\
    \midrule
    Teacher    & 32W32A      & 69.76     & 89.08 &73.30&91.42&76.13&92.86&68.79&88.68&71.88&90.29 \\
    Ours & 4W4A      & 69.42 & 88.91 &72.89&91.12&74.34&92.31&68.10&88.37&68.58&88.86\\
    Ours & 8W8A      & 69.58 & 89.11 &73.00&91.22&75.09&92.53&68.59&88.67&69.28&89.95\\
    \bottomrule
    \end{tabular}
\end{table}

When the transformation is not applied, we also applied our method at higher bit width. The results are appealing because we use only 30k images randomly sampled from the original training set, without ground truth.
\begin{table}[H]
    \scriptsize
    \caption{Imagenet performance of higher bit width, without transformations.}
    \label{results-on-imagenet-high-bit-wo_trans}
    \centering
    \begin{tabular}{cccccccccccc}
    \toprule
    \multirow{2}{*}{Method} & \multirow{2}*{Bit width}     & \multicolumn{2}{c}{Resnet18} & \multicolumn{2}{c}{Resnet34}& \multicolumn{2}{c}{Resnet50}& \multicolumn{2}{c}{Mobilenet V1}& \multicolumn{2}{c}{Mobilenet V2} \\
    \cmidrule(r){3-4}  \cmidrule(r){5-6} \cmidrule(r){7-8}  \cmidrule(r){9-10} \cmidrule(r){11-12}
    &&Top1 & Top5 & Top1 & Top5 & Top1 & Top5 & Top1 & Top5 & Top1 & Top5 \\
    \midrule
    Teacher    & 32W32A      & 69.76     & 89.08 &73.30&91.42&76.13&92.86&68.79&88.68&71.88&90.29 \\
    Ours & 4W4A      & 68.89 & 88.49 &72.50&90.87&75.53&92.61&65.44&86.92&69.14&88.86\\
    Ours & 8W8A      & 69.50 & 88.88 &73.08&91.20&75.84&92.81&68.36&88.61&71.46&90.23\\
    \bottomrule
    \end{tabular}
\end{table}

\subsection{How Does Concatenation + Convolution Affects?}
We provide an optional transformation of replacing the residual add with the Concatenation + Convolution architecture and we carry out experiments on ImageNet. What if we apply the other transformations and keep the residual add? We experiment on Resnet18. As shown in Table \ref{comparison-residual-add-concat-conv}, the quantization after the add operation brings a little loss of accuracy, while the replacement makes up for this loss, at the cost of the increase of computation expanse. On the whole, such a replacement is meaningful when an ASIC chip provides no native supports for the add operation.

\begin{table}[H]
    \caption{Comparison of the residual add and the concatenation + convolution.}
    \label{comparison-residual-add-concat-conv}
    \centering
    \begin{tabular}{cccc}
    \toprule
    \multirow{2}{*}{Method} & \multirow{2}*{Bit width}     & \multicolumn{2}{c}{Resnet18} \\
    \cmidrule(r){3-4} 
    &&Top1 & Top5 \\ 
    \midrule
    Teacher    & 32W32A      & 69.76     & 89.08 \\
    Residual Add & \multirow{2}{*}{2W4A}      & 67.90 & 88.14 \\
    Concat + Conv & & 68.77 & 88.66 \\
    \bottomrule
    \end{tabular}
\end{table}

\end{document}